%% file: paper.tex
\documentclass[nohyperref]{article}

\usepackage{microtype}
\usepackage{graphicx}
\usepackage{subfigure}
\usepackage{booktabs} % for professional tables
\usepackage[dvipsnames]{xcolor}
\usepackage{hyperref}

 \usepackage[accepted]{icml2022}

\usepackage{amsmath}
\usepackage{amssymb}
\usepackage{mathtools}
\usepackage{amsthm}

\usepackage{import}

\usepackage[capitalize,noabbrev]{cleveref}

\theoremstyle{plain}
\newtheorem{theorem}{Theorem}[section]
\newtheorem{proposition}[theorem]{Proposition}

\newtheorem{task}[theorem]{Task}
\theoremstyle{definition}

\theoremstyle{remark}

\usepackage[textsize=tiny]{todonotes}

\icmltitlerunning{Calibrated and Sharp Uncertainties in Deep Learning via Density Estimation}

\begin{document}

\twocolumn[
%\icmltitle{Submission and Formatting Instructions for \\
%           International Conference on Machine Learning (ICML 2022)}
\icmltitle{Calibrated and Sharp Uncertainties in Deep Learning via Density Estimation}

% It is OKAY to include author information, even for blind
% submissions: the style file will automatically remove it for you
% unless you've provided the [accepted] option to the icml2022
% package.

% List of affiliations: The first argument should be a (short)
% identifier you will use later to specify author affiliations
% Academic affiliations should list Department, University, City, Region, Country
% Industry affiliations should list Company, City, Region, Country

% You can specify symbols, otherwise they are numbered in order.
% Ideally, you should not use this facility. Affiliations will be numbered
% in order of appearance and this is the preferred way.
\icmlsetsymbol{equal}{*}

\begin{icmlauthorlist}
\icmlauthor{Volodymyr Kuleshov}{cornell_tech}
\icmlauthor{Shachi Deshpande}{cornell_tech}
%\icmlauthor{Firstname3 Lastname3}{comp}
%\icmlauthor{Firstname4 Lastname4}{sch}
%\icmlauthor{Firstname5 Lastname5}{yyy}
%\icmlauthor{Firstname6 Lastname6}{sch,yyy,comp}
%\icmlauthor{Firstname7 Lastname7}{comp}
%%\icmlauthor{}{sch}
%\icmlauthor{Firstname8 Lastname8}{sch}
%\icmlauthor{Firstname8 Lastname8}{yyy,comp}
%\icmlauthor{}{sch}
%\icmlauthor{}{sch}
\end{icmlauthorlist}

\icmlaffiliation{cornell_tech}{Department of Computer Science, Cornell Tech and Cornell University, New York, NY}
%\icmlaffiliation{comp}{Company Name, Location, Country}
%\icmlaffiliation{sch}{School of ZZZ, Institute of WWW, Location, Country}

\icmlcorrespondingauthor{Volodymyr Kuleshov}{kuleshov@cornell.edu}
\icmlcorrespondingauthor{Shachi Deshpande}{ssd86@cornell.edu}

% You may provide any keywords that you
% find helpful for describing your paper; these are used to populate
% the "keywords" metadata in the PDF but will not be shown in the document
%\icmlkeywords{Machine Learning, ICML}

\vskip 0.3in
]

% this must go after the closing bracket ] following \twocolumn[ ...

% This command actually creates the footnote in the first column
% listing the affiliations and the copyright notice.
% The command takes one argument, which is text to display at the start of the footnote.
% The \icmlEqualContribution command is standard text for equal contribution.
% Remove it (just {}) if you do not need this facility.

\printAffiliationsAndNotice{}  % leave blank if no need to mention equal contribution
%\printAffiliationsAndNotice{\icmlEqualContribution} % otherwise use the standard text.

\input{macros}

\newcommand\vk[1]{\textcolor{red}{[VK: #1]}}

\subimport{}{abstract}
\subimport{}{intro}

\subimport{}{background}

\subimport{}{algorithm}
\subimport{}{recalibration}
\subimport{}{calibration}

\subimport{}{experiments}

\subimport{}{discussion}

\paragraph{Acknowledgements}
This work is supported by a Tata Consulting Services grant on probabilistic deep learning.

\bibliographystyle{icml2022}
\bibliography{paper}

\end{document}

%% file: macros.tex
\providecommand{\tT}{\tilde T}
\providecommand{\var}{\textrm{var}}
\providecommand{\calib}{\textrm{cal}}
\providecommand{\sha}{\textrm{sha}}

\providecommand{\x}{\chi}
\providecommand{\e}{\epsilon}
\providecommand{\La}{\Lambda}
\providecommand{\la}{\lambda}
\providecommand{\vt}{\tilde{v}}
\providecommand{\ut}{\tilde{u}}
\providecommand{\pit}{\tilde{\pi}}
\providecommand{\Vt}{\tilde{V}}
\providecommand{\Ut}{\tilde{U}}
\newcommand{\supr}[1]{^{(#1)}}

\providecommand{\sMh}{\widehat{\sM}}
\providecommand{\Mh}{\widehat{M}}
\providecommand{\Th}{\widehat{T}}
\providecommand{\vh}{\widehat{v}}

\providecommand{\pb}{\bar{p}}
\providecommand{\pib}{\bar{\pi}}
\providecommand{\Ub}{\bar{U}}
\providecommand{\Vb}{\bar{V}}
\providecommand{\Lb}{\bar{\Lambda}}

\providecommand{\off}{\operatorname{off}}

\newcommand{\yh}{{\hat y}}
\newcommand{\ellavg}{{\bar \ell}}
\newcommand{\Yc}{{\mathcal Y}}
\newcommand{\Ic}{{\mathcal I}}
\newcommand{\Bcal}{{\mathcal B}}
\newcommand{\Fc}{{\mathcal F}}
\newcommand{\Xc}{{\mathcal X}}
\newcommand{\Pc}{{\mathcal P}}
\newcommand{\Rb}{{\mathbb R}}
\newcommand{\extR}{R^\mathrm{ext}}
\newcommand{\intR}{R^\mathrm{int}}
\newcommand{\Fcal}{F^\mathrm{cal}}
\newcommand{\Ind}{{\mathbb{I}}}
\newcommand{\Exp}{{\mathbb{E}}}
\newcommand{\Pb}{{\mathbb{P}}}
\newcommand{\palg}{p^F}
\newcommand{\wsupj}{\Ind^{(j)}}
\newcommand{\ren}{{\rho^\e_n}}
\newcommand{\rjt}{{\rho^{(j)}_T}}
\newcommand{\wi}{w^{(i)}}
\newcommand{\bfw}{{\bf w}}
\newcommand{\bfd}{{\bf d}}
\newcommand{\bfp}{{\bf p}}
\newcommand{\pk}{p^{(k)}}
\newcommand{\Regret}{\textrm{Regret}}

\newtheorem{claim}{Claim}
\newtheorem{defn}{Definition}
\newtheorem{thm}{Theorem}

%% file: abstract.tex
\begin{abstract}

Accurate probabilistic predictions can be characterized by two properties—calibration and sharpness. However, standard maximum likelihood training yields models that are poorly calibrated and thus inaccurate—a 90\% confidence interval typically does not contain the true outcome 90\% of the time. 
This paper argues that calibration is important in practice and is easy to maintain by performing low-dimensional density estimation. 
We introduce a simple training procedure based on recalibration that yields calibrated models without sacrificing overall performance; unlike previous approaches, ours ensures the most general property of distribution calibration and applies to any model, including neural networks.
We formally prove the correctness of our procedure assuming that we can estimate densities in low dimensions and we establish uniform convergence bounds. 
Our results yield empirical performance improvements on linear and deep Bayesian models and suggest that calibration should be increasingly leveraged across machine learning. We release a library$^2$ that implements our methods along with a blog post here:

{\color{Orange}
https://shachideshpande.github.io/blog-distribution-calibration/}

\end{abstract}

%% file: intro.tex
\begin{figure}[t]
%\begin{center}
\hspace{-5mm}\includegraphics[width=9cm]{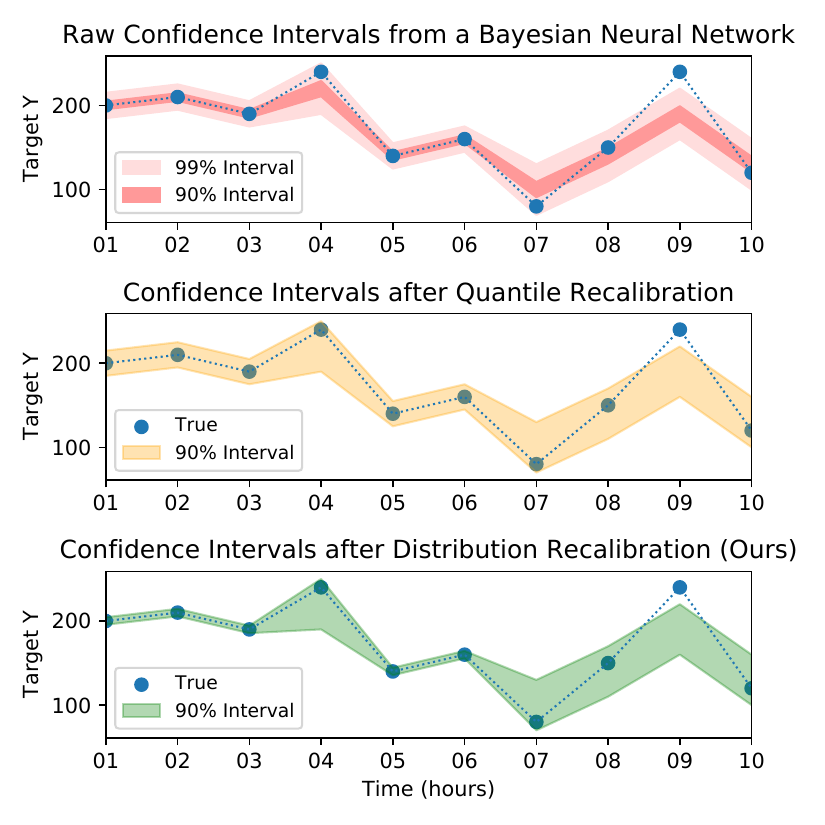}
%\end{center}
\vspace{-9mm}
%\caption{Calibration of Bayesian neural networks. Top: A Bayesian neural network outputs a probability distribution over a continuous variable (blue), from which we derive a 90\% confidence interval (red). This confidence interval fails to accurately capture mode uncertainty: most points fall outside. Bottom: Our method recalibrates the neural network and outputs a 90\% confidence interval (green) that correctly contains 9 out of 10 points.  }
\label{fig:first}
\caption{Top: Credible intervals (90\% \& 99\%) from a Bayesian neural network on a time series forecasting task. The 90\% credible interval around the forecast (red) is miscalibrated and inaccurate: half of the points are outside of it. Middle: Quantile recalibration \citep{kuleshov2018accurate} relabels the 99\% credible interval as a 90\% interval, which now correctly contains 9/10 points (orange). Bottom: Our recalibration method enforces distribution calibration---the 90\% interval is unchanged at points where it is narrow and expands where it is wide. This yields an improved and narrower calibrated 90\% interval (green).}
\vspace{-2mm}
\end{figure}

\section{Introduction}\label{sec:introduction}

Accurate probabilistic predictions can be characterized by two properties---calibration and sharpness \citep{gneiting2007probabilistic}. Intuitively, calibration means that a 90\% confidence interval contains the true outcome 90\% of the time. Sharpness means that these confidence intervals are narrow.
These properties are grounded in statistical theory and are used to evaluate forecasts in domains such as meteorology and medicine \citep{gneiting2005weather,gneiting2007strictly}.

This paper argues for reasoning about uncertainty in deep learning in terms of calibration and sharpness and proposes simple algorithms for enforcing these properties.
Standard maximum likelihood training yields models that are poorly calibrated—a 90\% confidence interval typically does not contain the true outcome 90\% of the time \citep{guo2017calibration}. 
Popular approaches based on dropout or ensembling \citep{Gal2016Dropout,lakshminarayanan2016simple} improve model probabilities, yet do not enforce calibration \citep{kuleshov2018accurate}.
Recalibration techniques \citep{platt1999probabilistic} yield calibrated and sharp forecasts, but enforce a limited notion of quantile calibration \citep{kuleshov2018accurate}, require complex variational approximations \citep{song2019distribution}, and are not well understood theoretically.

This paper argues that calibration is important in practice and is easy to maintain by performing low-dimensional density estimation. 
We introduce a simple recalibration procedure based on quantile function regression \citep{si2021autoregressive}, which yields calibrated models without sacrificing overall performance.
Unlike previous approaches, ours guarantees the general property of distribution calibration \citep{song2019distribution} and applies to any model, including neural networks.
We prove the correctness of our procedure assuming that we can estimate densities in low dimensions and we establish uniform convergence bounds.

Empirically, we find that our method consistently outputs well-calibrated predictions in linear and deep Bayesian models, and improves performance on downstream tasks with minimal implementation overhead.
A key takeaway is that distribution calibration may be simpler to obtain than previously thought, and we argue that it should be enforced in predictive models and leveraged across machine learning.

\paragraph{Contributions.}
In summary, we make three contributions.
We show that probability calibration (including distribution calibration) can be maintained via density estimation in low dimensions. We then introduce a simple algorithm based on recalibration that is simpler and more broadly applicable than previous approaches. We complement this algorithm with a theoretical analysis establishing guarantees on calibration and vanishing regret.
A key takeaway is that calibration may be easier to maintain than previously thought and should be leveraged more broadly throughout machine learning.

 \begin{figure*}[t]
\begin{center}
 \includegraphics[width=16.5cm]{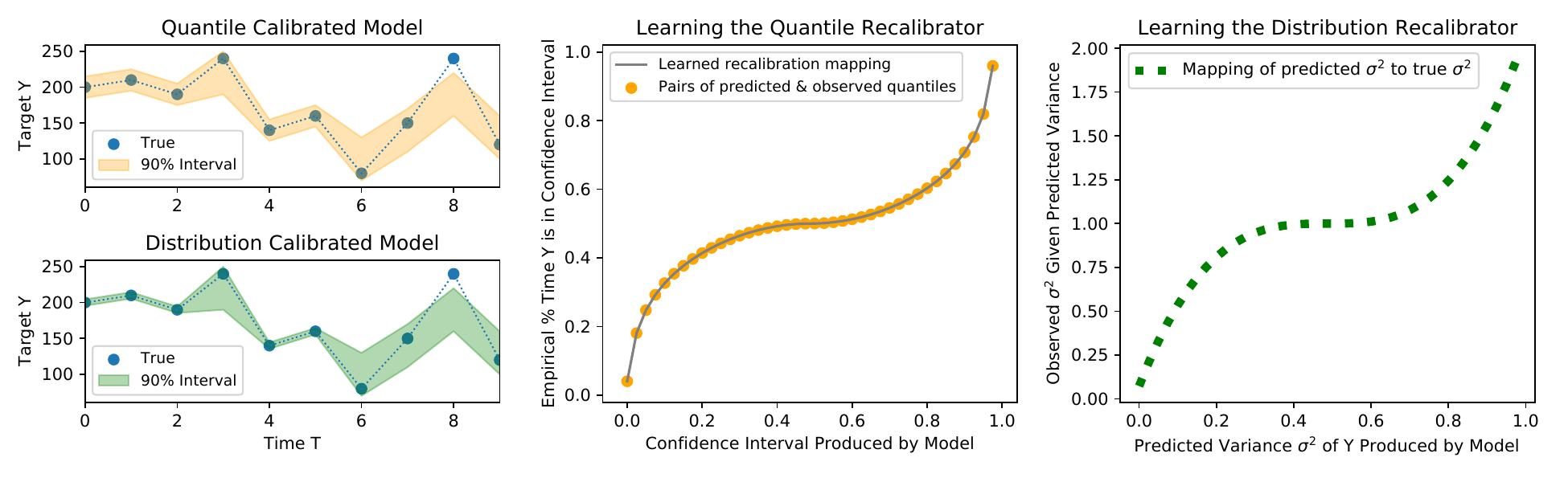}
\end{center}
\vspace{-4mm}
\caption{Quantile vs. distribution calibration.. 
Left: A Bayesian neural network outputs probabilistic forecasts recalibrated using quantile and distribution recalibration.
Middle: In quantile recalibration, we observe each confidence interval and count how many times $y$ falls in that interval---this mapping is learned by $R$ and is an estimate of $\mathbb{P}(F_X(Y) \leq p)$.
Right: In distribution recalibration, we represent forecasts by features (e.g., their variance $\sigma^2$) and learn a density $\mathbb{P}(Y \mid F)$ conditioned on these features. In this example, we learn to remap predicted to observed variance---this allows adjusting different variances differently, yielding tighter intervals (bottom left).
}\label{fig:regression}
\end{figure*}

%% file: background.tex
\section{Background}\label{sec:background}

% We start with a background on predictive uncertainties, presenting it in a way that will be useful later in the paper. % VK: Need to explain this??
% Machine learning models often output probabilistic predictive uncertainties.
% This section provides a background on probabilistic predictive modeling 
% and summarizes this topic from a unified perspective that will make be useful for establishing our later results.
% %and offers a reinterpretation of existing techniques that will be useful for deriving an extension to the regression and Bayesian settings in the next section.

\subsection{Predictive Uncertainty in Machine Learning}

Supervised machine learning models predict a probability distribution over the target variable---e.g. class membership probabilities or the parameters of a normal distribution. 
We seek to produce models with accurate probabilistic outputs.
%Such distributions are useful for interpretability, safety, and downstream decision-making.
%
%Aleatoric uncertainty captures the inherent noise in the data, while epistemic uncertainty arises from not having a large enough dataset to estimate model parameters \citep{kendall2017uncertainties}.
% Our work applies to both types of uncertainty.
% The results in this paper apply to both types of uncertainties.
% The results in this paper apply equally to aleatoric uncertainties as well as to epistemic uncertainties obtained using a Bayesian framework.

\paragraph{Notation.}

%Supervised models predict a target $y \in \mathcal{Y}$ from an input $x \in \mathcal{X}$.
%, where $x, y$ are realizations of random variables $X, Y \sim \mathbb{P}$ and  $\mathbb P$ is the data distribution.
We predict a target  $y \in \mathcal{Y}$---where $\mathcal{Y}$ is either discrete (it's a classification problem) or $\mathcal{Y} = \mathbb{R}$ (it's a regression problem)---using input features $x \in \mathcal{X}$.
%The outputs $y \in \mathcal{Y}$ either take one of $K$ discrete values, which yields a classification problem, or $\mathcal{Y} = \mathbb{R}$, which yields regression.
We are given
a forecaster $H : \mathcal{X} \to \Delta_\mathcal{Y}$, which outputs a probability distribution $F(y) : \mathcal{Y} \to [0,1]$ within the set $\Delta_\mathcal{Y}$ of distributions over $\mathcal{Y}$; the probability density function of $F$ is $f$.
We are also given a training set $\mathcal{D} = \{ (x_i, y_i) \in \mathcal{X} \times \mathcal{Y} \}_{i=1}^n$ and a calibration set $\mathcal{C} = \{ (x_j, y_j) \in \mathcal{X} \times \mathcal{Y} \}_{j=1}^m$, each
consisting of i.i.d.~realizations of  random variables $X, Y \sim \mathbb{P}$, where $\mathbb P$ is the  data distribution.
%Both sets contain i.i.d.~realizations of random variables $X, Y \sim \mathbb{P}$, where $\mathbb P$ is the  data distribution.

%Formally, we are given
%%We are interested in probabilistic classification and regression settings, in which
%%Formally, we say that 
%a forecaster $H : \mathcal{X} \to \Delta_\mathcal{Y}$, which outputs a probability distribution $F(y) : \mathcal{Y} \to [0,1]$ within the set $\Delta_\mathcal{Y}$ of distributions over $y$. 
%%The output space $\mathcal{Y}$ takes one of $K$ possible discrete values, we have a classification problem
%The $y \in \mathcal{Y}$ are scalar-valued discrete or continuous variables; 
%we use $f$ to denote the probability density function associated with $F$.
%%We use $f$ to denote the probability density or probability mass function associated with $F$.
%% targeting the label $y$. 
%% In this paper, we assume that $F$ is a probability mass or a probability density function, depending on the form of $y$.
%%
%We work with a training set $\mathcal{D} = \{ (x_i, y_i) \in \mathcal{X} \times \mathcal{Y} \}_{i=1}^n$ and a calibration set $\mathcal{C} = \{ (x_j, y_j) \in \mathcal{X} \times \mathcal{Y} \}_{j=1}^m$.
%Both sets contain i.i.d.~realizations of random variables $X, Y \sim \mathbb{P}$, where $\mathbb P$ is the  data distribution.
%%The model $H$ is trained on a labeled dataset $x_i, y_i\in \mathcal{X} \times \mathcal{Y}$ for $i=1,2,...,n$ of  i.i.d.~realizations of random variables $X, Y \sim \mathbb{P}$, where $\mathbb P$ is the  data distribution.

% TODO: Verify formulas!
\begin{table*}
\begin{center}
\begin{tabular}{l|c|c|c}
\toprule
{\bf Proper Loss \hspace{5mm}} & {\bf Loss} & {\bf Calibration} & {\bf Refinement} \\
& $L(F,G)$ & $L_c(F,Q)$ & $L_r(Q)$ \\
\midrule
Logarithmic & $\Exp_{y\sim G}$ $\log f(y)$ & $\text{KL}(q||f)$ & $H(q)$ \\
CRPS & \hspace{2mm} {$\Exp_{y\sim G}$ $(F(y) - G(y))^2$} \hspace{2mm}& {$\int^{\infty}_{-\infty}(F(y) - Q(y))^2$dy} & \hspace{2mm} {$\int^{\infty}_{-\infty} Q(y) (1 - Q(y))dy$} \\
Quantile & {$\Exp^{\tau\in U[0,1]}_{y\sim G} \rho_\tau(F, y)$} & \hspace{2mm} {$\int_0^1 \int^{F^{-1}(\tau)}_{Q^{-1}(\tau)}(Q(y) - \tau)dyd\tau$} \hspace{2mm} & {$\Exp^{\tau\in U[0,1]}_{y\sim Q} \rho_\tau(Q, y)$} \\
\bottomrule
\end{tabular}
\end{center}
\caption{Examples of three proper losses: the log-loss, the continuous ranked probability score (CRPS), and the quantile loss. A proper loss $L(F,G)$ between distributions $F, G$---assumed here to be cumulative distribution functions (CDFs)---decomposes into a calibration loss term $L_c(F,Q)$ (also known as reliability) plus a refinement term $L_r(Q)$ (which itself decomposes into a sharpness and an uncertainty term). Here, $Q(y)$ denotes the CDF of $\mathbb{P}(Y=y \mid F_X = F)$, and $q(y), f(y)$ are the probability density functions of $Q$ and $F$.}\label{tbl:properlosses}
\end{table*}

% % TODO: Verify formulas!
% \begin{table}
% \begin{center}
% \begin{tabular}{c|c|c|c}
% \toprule
% {\bf Proper Loss} & {\bf Calibration} & {\bf Sharpness} & {\bf Uncertainty} \\
% $L(F,G)$ & $L_c(F,Q)$ & $L_s(Q,\bar Q)$ & $\sigma(\bar Q)$ \\
% \midrule
% $\Exp_{y\sim G}$ $\log F(y)$ & $\text{KL}(Q||F)$ & $\text{KL}(Q||\bar Q)$ & $H(\bar Q)$ \\
% {\scriptsize $\Exp_{y\sim G}$ $(F(y) - G(y))^2$} & {\scriptsize $\sum_y(F(y) - Q(y))^2$} & {\scriptsize $\sum_y(Q(y) - \bar Q(y))^2$} & {\scriptsize $\sum_y \bar Q(y) (1 - \bar Q(y))$} \\
% \bottomrule
% \end{tabular}
% \end{center}
% \caption{Examples of proper loss functions. A proper loss is a function $L(F,G)$ over a forecast $F$ targeting a variable $y \in \mathcal{Y}$ whose true distribution G. Each $L(F,G)$ decomposes into a sum $L_c(F,Q) - L_s(Q,\bar Q) + \sigma(\bar Q)$ of three terms: calibration (also known as reliability), sharpness, and an irreducible uncertainty term. These involve the distribution $Q(y) := \mathbb{P}(Y=y \mid F_X = F)$ and $\bar Q(y) = \mathbb{P}(Y=y)$. }\label{tbl:properlosses}
% \end{table}

% \subsection{Calibration and Sharpness}
\subsection{Calibration and Sharpness}
%\subsection{Calibration and Sharpness---Two Qualities of an Ideal Prediction}
% \subsection{Ideal Uncertainties are Calibrated and Sharp}

%We seek to produce predictive forecasts that are {\em sharp} and {\em calibrated}. Calibration admits several definitions.
%Predictive probabilities can be characterized by two properties---calibration and sharpness. We introduce these next.
Accurate probabilistic predictions are characterized by two properties---calibration and sharpness.
Calibration means that an $80\%$ confidence interval contains the target $y$ $80\%$ of the time. Sharpness means that the intervals are tight.

\paragraph{Quantile Calibration in Regression}

%\citet{kuleshov2018accurate} propose algorithms for
%A closely related, but weaker concept is 
%quantile calibration, which asks that a 90\% confidence interval around a prediction contains the true target $y$ 90\% of the time. It can be defined as:
Formally, \citet{kuleshov2018accurate} define calibration in a regression setting as
\begin{equation}
\mathbb{P}(Y \leq \text{CDF}_{F_X}^{-1}(p)) = p \; \textrm{for all $p \in [0,1]$}, \label{eqn:calibration0}
% \label{eqn:calibration2}
\end{equation}
where $F_X = H(X)$ is the forecast at $X$, itself a random variable that takes values in $\Delta_\mathcal{Y}$.
Intuitively, for each $X, Y$, the $Y$ is contained in the $p$-th confidence interval $(-\infty, \text{CDF}_{F_X}^{-1}(p)]$ predicted at $X$ a fraction $p$ of the time.
%Quantile calibration is implied by distributional calibration \citep{song2019distribution}.
%\citet{kuleshov2018accurate} introduce a recalibration-based approach, in which a model $R : [0,1] \to [0,1]$ is fit to approximate $\mathbb{P}(Y \leq \text{CDF}_{F_X}^{-1}(p))$; as a result, the forecasts $R \circ F$ are calibrated.

% In binary classification, calibration intuitively means that of the times when we predicted that $y=1$ with probability $p$, the event $y=1$ should occur 80\% of the time. %we made that prediction.
% Intuitively, calibration means that if we say that the probability of rain is 80\%, then it should rain 80\% of the times we made that prediction.
% $y_t$ should fall in a 90\% confidence interval approximately 90\% of the time. % in practice.
%

\paragraph{Distribution Calibration}
\citet{song2019distribution} defines a stronger notion of distribution calibration as
%Formally, calibration can be defined by the equation
\begin{equation}
\mathbb{P}(Y = y \mid F_X = F) = f(y)
\textrm{ for all $y \in \mathcal{Y}$, $F \in \Delta_{\mathcal{Y}}$}, \label{eqn:calibration1}
\end{equation}
where $F_X = H(X)$ is the random forecast at $X$ and $f$ is its probability density or probability mass function.
%where $X, Y \sim \mathbb{P}$ are random variables corresponding to the input features and targets, and  
%$F_X = H(X)$ is the random forecast at $X$. %, itself a random variable that takes values in $\Delta_\mathcal{Y}$. 
%We use $f$ to denote the probability density or probability mass function associated with $F$.
%
When $\mathcal{Y}=\{0,1\}$ and $F_X$ is Bernoulli with parameter $p$, we can write \eqref{eqn:calibration1} as $\mathbb{P}(Y = 1 \mid F_X = p) = p$. Intuitively, the true probability of $Y=1$ is $p$ conditioned on predicting it as $p$.

Equation \ref{eqn:calibration1} extends to regression as well. For example, if $F$ is a Gaussian with variance $\sigma^2$, this definition asks that the data distribution conditioned on predicting $F$ also has variance $\sigma^2$. 
Distribution calibration also implies that quantile calibration holds \citep{song2019distribution}.
%Distribution calibration is a stronger concept that implies quantile calibration \citep{song2019distribution}.
%This recently proposed definition is called {\em distribution} calibration \citep{song2019distribution}, and it implies 

% When the $x_t, y_t$ are i.i.d.~realizations of random variables $X, Y \sim \mathbb{P}$, a sufficient condition for this is
% \begin{align}
% \mathbb{P}(Y \leq F_X^{-1}(p)) = p \; \textrm{for all $p \in [0,1]$}, \label{eqn:calibration2}
% \end{align}
% %{\color{red}[SE: again, why approx here? seems sloppy]}
% where we use $F_X = H(X)$ to denote the forecast at $X$. This formulation is related to the notion of probabilistic calibration of \citet{gneiting2007probabilistic}.%; see \sectionref{discussion} for more details.\vk{need this?}

\paragraph{Calibration vs.~Sharpness}
Calibration by itself is not sufficient to produce a useful forecast. For example, a binary classifier that always predicts the marginal data probability $\mathbb{P}(Y = 1)$ is calibrated; however it does not even use the features $X$ and thus cannot be accurate. %{\color{red}[SE: might want to precisely define the forecastER as well to be more precise]}

In order to be useful, forecasts must also be {\em sharp}. Intuitively, this means that the confidence intervals should be as tight as possible.
Formally, sharpness is quantified via the entropy of $F_X$; see Table \ref{tbl:properlosses} and the 
%More formally, we want the entropy of the predicted probabilities to be small. 
%This is captured by proper scoring rules as part of a refinement term (see Table \ref{tbl:properlosses}), which equals an irreducible term minus a sharpness term \citep{murphy1973vector,brocker2009decomposition}. The latter is maximized when we minimize the scoring rule.

\subsection{Evaluating Probabilistic Predictions} 

% It is usually easy to compare point estimates from supervised learning models using metrics such as accuracy or mean squared error.
% Predictive probabilities, on the other hand, are more complex and require specialized metrics.
% While point estimates from supervised learning models are usually easy to compare  using metrics such as accuracy or mean squared error, predictive probabilities are more complex, and require specialized tools.

We evaluate probabilistic predictions using the framework of proper scoring rules \citep{gneiting2007strictly}. 
%In statistics, the standard tool for evaluating the quality of predictive forecasts is a proper scoring rule \citep{gneiting2007strictly}. 
%This paper advocates for evaluating the quality of predictive uncertainties using proper scoring rules \citep{gneiting2007strictly}. 
Formally, let $L : \Delta_\mathcal{Y} \times \mathcal{Y} \to \mathbb{R}$ denote a loss between a probabilistic forecast $F \in \Delta_\mathcal{Y}$ and a realized outcome 
$y \in \mathcal{Y}$. Given a distribution $G \in \Delta_\mathcal{Y}$ over $y$, we use $L(F,G)$ to denote the expected loss
$
L(F,G) = \mathbb{E}_{y \sim G} L(F, y).
$

We say that $L$ is a {\em proper loss} if it is minimized by $G$ when $G$ is the true distribution for $y$:
$
L(F,G) \geq L(G,G) \text{ for all $F$}.
$
%When $S$ is proper, we also refer to it as a proper loss.
One example is the log-likelihood $L(F,y) = - \log f(y)$. 
%where $f$ is the probability density or probability mass function of $F$. 
Another example is the check score for $\tau \in [0,1]$:
%\begin{equation}
%\rho_\tau(F, y) = \tau (y-F^{-1}(\tau)) \text{ if $y \geq f$ else $(1-\tau)(F^{-1}(\tau)-y)$}
%\end{equation}
\begin{equation}\label{eqn:checkscore}
\rho_\tau(F, y) = 
\begin{cases}
\tau (y-F^{-1}(\tau)) & \text{ if $y \geq f$} \\
(1-\tau)(F^{-1}(\tau)-y) & \text{ otherwise}.
\end{cases}
\end{equation}
%Another common loss is the check score $\rho_\tau(y, f) = \tau (y-f)$ if $y \geq f$ and $(1-\tau)(f-y)$ otherwise; it can be used to estimate the $\tau$-th quantile of a distribution.
See Table \ref{tbl:properlosses} for additional examples.

% \paragraph{Calibration and Sharpness.}
%What are the qualities of a good probabilistic prediction, as measured by a proper scoring rule?
%It can be shown that 
Every proper loss decomposes into a sum of the following terms \citep{gneiting2007probabilistic}:
$$\text{proper loss} = \text{calibration} \underbrace{- \text{sharpness} + \text{irreducible term}}_\text{refinement term}.$$
Thus, calibration and sharpness are both necessary and sufficient characteristics of an accurate probabilistic forecast.
%Thus, there are precisely two qualities that define an ideal forecast: calibration and sharpness.
% These represent precisely the two desiderata that matter for evaluating uncertainties. 
%We examine each of them next.

\subsection{Training Calibrated Models}
%\subsection{Training Models to Output Accurate Probabilities}

Typically, machine learning models do not output accurate and calibrated probabilities out-of-the-box \citep{niculescu2005predicting,guo2017calibration}. Recalibration is an alternative training strategy
%\citet{kuleshov2018accurate} introduce a recalibration-based approach, 
in which a model $R : \Delta_\mathcal{Y} \to \Delta_\mathcal{Y}$ is fit on the calibration set $\mathcal{C}$ such that the forecasts $R \circ F$ are calibrated \citep{platt1999probabilistic,vovk2005algorithmic}

In the setting of quantile calibration, one can see that choosing $R : [0,1] \to [0,1]$ to be $R(p) = \text{CDF}_{F_X}^{-1}(p))$ yields a calibrated $R \circ F$ as in (\ref{eqn:calibration0}). Thus, recalibration can be understood as estimating the above distribution.
%to approximate $\mathbb{P}(Y \leq \text{CDF}_{F_X}^{-1}(p))$; as a result, the forecasts $R \circ F$ are calibrated.

%% file: algorithm.tex
% \section{Obtaining Calibrated And Sharp Uncertainties in Practice}\label{sec:algorithm}
%\section{Distribution Calibration via Density Estimation}\label{sec:algorithm}
\section{Training Distribution Calibrated Models}\label{sec:algorithm}

%This  section  introduces  algorithms  that  ensure  the  distribution calibration of a predictive model. 
%Unlike existing methods for distribution calibration, ours can be used with any forecaster (not just ones that output Gaussians), are very simple to implement in differentiable programming frameworks, and have theoretical guarantees.

In this section, we introduce methods for training probabilistic classification and regression models that ensure the strong property of distribution calibration. 
Our approach is an instance of recalibration \citep{platt1999probabilistic,kuleshov2018accurate}, in which we learn an auxiliary model $R : \Delta_\mathcal{Y} \to \Delta_\mathcal{Y}$ such that the joint model $R \circ H$ is calibrated.

\subsection{Recalibration as Density Estimation}

When $x_t, y_t$ are sampled i.i.d.~from $\mathbb{P}$, choosing
\begin{equation}
R(F) = \mathbb{P}(Y \mid H(X)=F)\label{eqn:density}
\end{equation}
yields a distribution calibrated model $R \circ H$; we provide a formal proof in Section \ref{sec:theory}.
Thus, our task is to learn the density (\ref{eqn:density}); this requires solving two challenges: (1) conditioning $R$ on arbitrary forecasts $F$; (2) choosing a learning objective for $R$.

%\paragraph{Representing Distributions}
\paragraph{Challenge 1: Conditioning on $F$}

Our approach is to define a featurization $\phi : \Delta_\mathcal{Y} \to \mathbb{R}^p$ of $F$, such that $\phi(F)$ is represented by a small number of parameters $p$.
%Estimating the distribution in (\ref{eqn:density}) is challenging because we are conditioning on a general function $F \in \Delta_\mathcal{Y}$. 
Hence, learning $\mathbb{P}(Y \mid \phi(F))$ involves a tractable low-dimensional density estimation problem for which there exist efficient and provably correct algorithms \citep{wasserman2006all}.

In practice, most models (especially neural networks) already assume a parametric form for $F$ (e.g., $\mu, \sigma^2$ for a Gaussian). Alternatively, we may featurize $F$ via some of its quantiles. We provide several additional examples of features in Section \ref{sec:qfr}.

%In practice, most models (especially neural networks) will represent $F$ via parameters $\phi_F$ within a set $\Phi_F(\Delta_\mathcal{Y})$. For example, it is common to assume that $F$ is Gaussian, in which case $\phi_F = (\mu, \sigma^2)$. Similarly, we assume the output forecast $G = R(F)$ is described by parameters $\phi_G$ in $\Phi_G (\Delta_\mathcal{Y})$.

%In any case, when $Y$ is scalar-valued, we can represent $F$ as a low-dimensional vector. Hence, learning $\mathbb{P}(Y \mid H(X)=F)$ involves a low-dimensional density estimation problem that will be tractable and admits mature algorithms (many of which have theoretical guarantees).

%\paragraph{The Structure of a Recalibration Method}
%
%At a high level, a recalibration method takes a pre-trained model $H$ and performs two steps: 
%\begin{enumerate}
%\item Running $H$ on a calibration set $\mathcal{C} = \{x_t,y_t\}_{t=1}^m$ sampled i.i.d.~from $\mathbb{P}$ to obtain a dataset of model outputs $\mathcal{S} = \{ (H(x), y) \mid x, y \in \mathcal{C}\}$; 
%\item Training a model $R : \Delta_\mathcal{Y} \to \Delta_\mathcal{Y}$ over $S$. 
%\end{enumerate}
%This requires choosing a parameterization for $H(x)$ and a learning objective for $R$.

%\paragraph{Learning Densities Using Proper Scoring Rules}
%\paragraph{Density Estimation with Proper Scoring Rules}
\paragraph{Challenge 2: A Learning Objective for $R$}

%It is common to estimate densities via maximum likelihood estimation. 
%We extend this approach to use a general proper scoring rule $L(\cdot, \cdot)$, of which the log-likelihood is one special case.
%Estimating the distribution in (\ref{eqn:density}) also requires defining a learning objective.
Our approach will optimize a proper scoring rule $L$. %; maximum likelihood density estimation is a special case of this procedure when $L$ is the log-loss.
%We learn this density by optimizing a proper scoring rule $S(\cdot, \cdot)$. We give examples of specific scoring rules below. 
Specifically, we choose a recalibrator $R$ that minimizes the objective $\sum_{x, y \in \mathcal{C}} L(R(F_x), y)$ over a calibration dataset $\mathcal{C} = \{x_j,y_j\}_{j=1}^m$ sampled i.i.d.~from $\mathbb{P}$:. Observe that
\begin{align*}
\sum_{x, y \in \mathcal{C}} L(R(F_x), y) 
& \approx \Exp_{F_X \sim \mathbb{P}} \Exp_{Y \sim \mathbb{P}(Y|F_X)} [ L(R(F_X), Y) ] \\
& = \Exp_{F_X \sim \mathbb{P}} [ L(R(F_X), \mathbb{P}(Y|F_X)) ],
\end{align*}
where $F_X = H(X)$. The first line follows by Monte Carlo approximation and the second line follows from the definition of a proper loss. 

Thus, minimizing $L(R(F_x)), y)$ over $\mathcal{C}$ yields an estimate of $\mathbb{P}(Y|F_X)) = \mathbb{P}(Y|H(X) = F))$, which is the probability we seek to learn.
Overall, our high-level approach can be summarized by the following algorithm; we describe specific instantiations of $L, \phi$, and $R$ below.
\begin{algorithm}
  \caption{Distribution Recalibration Framework}
  \label{alg:recal0}
  %\begin{algorithmic}
  \textbf{Input:}
    Pre-trained model $H : \mathcal{X} \to \Delta_\mathcal{Y}$, featurizer $\phi : \Delta_\mathcal{Y} \to \mathbb{R}^p$, recalibrator $R : \mathbb{R}^p \to \Delta_\mathcal{Y}$,  calibration set $\mathcal{C}$\\
  \textbf{Output:}
    Recalibrated model $R \circ H : \mathcal{X} \to \Delta_\mathcal{Y}$ %\Delta_\mathcal{Y}$.
   \begin{enumerate}
%    \item Fit the base model on $\mathcal{D}$: $\min_{H} \sum_{(x, y) \in \mathcal{D}} L(H(x), y)$
    \item Create a training set for recalibrator: \\$\mathcal{S} = \{ (\phi(H(x)), y) \mid x, y \in \mathcal{C}\}$
    \item Fit the recalibrator $R$ on $\mathcal{S}$ using a proper loss: \\
    $\min_{R} \sum_{(\phi,y) \in \mathcal{S}} L\left(R(\phi), y\right)$ 
%    $\min_{R} \sum_{(\phi,y) \in \mathcal{S}} \Exp_{\tau\in U[0,1]}\rho_\tau(R^{-1}(\tau;\phi),y)$
   \end{enumerate}
\end{algorithm}

%\subsection{Practical Implementation}
%
%\paragraph{Notation}
%\vk{Use $\Phi_1(\Delta_\mathcal{Y})$ and mention $\Phi$ featurization in b/g}
%We first assume there exists a parameterization $\Phi_1$ of the probabilities returned by forecaster $H$: for each $p \in \Delta(\mathcal{Y})$ returned by $H$, there exist parameters $\phi \in \Phi_1$ that describe $p$. 
%The $\phi$ can be the natural parameters of an exponential family distribution, such as $(\mu, \sigma^2)$ describing a Gaussian.
%% In the simplest case, $\phi$ can be $(\mu, \sigma^2)$ describing a Gaussian; in more complex settings, $\phi$ could be the weights of a deep generative model over high-dimensional $\phi$.
%% We similarly assume a parameterization $\Phi_2$ of the outputs of $R$
%
%First, we train a base forecaster $H$ to minimize a proper loss $L$. Then, we train an auxiliary model $R : \Phi_1 \to \Phi_2$ (called the recalibrator) over the outputs of $H$ that outputs the parameters $\phi_2 \in \Phi_2$ of another distribution such that $L$ is minimized. Here $\Phi_2$ is a second parameterization of $\Delta(\mathcal{Y})$ (possibly the same). As a result, the forecasts $(R \circ H)(X)$ will be calibrated. We provide details in Algorithm \ref{alg:recal}.

%\subsection{Estimating a Cumulative Distribution Function}
%\subsection{Distribution Calibrated Regression via Quantile Function Estimation}\label{sec:qfr}
\subsection{Density Estimation via Quantile Function Regression}\label{sec:qfr}

Next, we introduce specialized versions of Algorithm \ref{alg:recal0} for the settings of probabilistic regression and classification and we define additional details of the method---the model $R$, the features $\phi$, and the objective $L$. %\vk{repetitive?}
%These methods make additional modeling assumptions that we describe below.

Recall that our goal is to estimate the distribution $\mathbb{P}(Y \mid H(X)=F)$ (Equation \ref{eqn:density}). We choose to represent this distribution via its cumulative distribution function (CDF)  or, equivalently, its inverse, the quantile function (QF).
Learning a model of the CDF or of the QF facilitates computing confidence intervals and yields more numerically stable algorithms.

Without loss of generality, we train $R$ to fit the QF; the density can be obtained from the CDF or QF by via a derivative. This approach also yields the following equivalent notion of distribution calibration:
%
%An alternative is to learn a model of the cumulative distribution function (CDF) of the distribution $\mathbb{P}(Y \leq y \mid F_X = F)$. This model corresponds to an equivalent notion of distribution calibration:
\begin{equation*}
\mathbb{P}(Y \leq y \mid F_X = F) = F(Y \leq y)
\textrm{ for all $y \in \mathcal{Y}$, $F \in \Delta_{\mathcal{Y}}$}, \label{eqn:calibration3}
\end{equation*}

%Introducing a regression version still requires defining a few properties---X, Y, and Z.

%We first propose an algorithm for distribution calibrated regression. When $Y$ is continuous, the distribution $\mathbb{P}(Y \mid H(X)=F)$ can be represented either via its density or via its cumulative distribution function (CDF). Unlike previous approaches, we model the CDF or, equivalently, its inverse, the quantile function (QF). Our approach will yield the following equivalent notion of distribution calibration:
%%
%%An alternative is to learn a model of the cumulative distribution function (CDF) of the distribution $\mathbb{P}(Y \leq y \mid F_X = F)$. This model corresponds to an equivalent notion of distribution calibration:
%\begin{equation*}
%\mathbb{P}(Y \leq y \mid F_X = F) = F(Y \leq y)
%\textrm{ for all $y \in \mathcal{Y}$, $F \in \Delta_{\mathcal{Y}}$}, \label{eqn:calibration3}
%\end{equation*}
%Learning a model of the CDF or of the QF enables easily computing confidence intervals and in our experience yields more numerically stable algorithms.
%%We outline our method in Algorithm \ref{alg:recal}.

Our approach for estimating the QF relies on quantile function regression \citep{si2021autoregressive}. We define this method below in terms of three components---the model $R$, the features $\phi$, and the objective $L$.
%We learn an estimate of $\mathbb{P}(Y \mid H(X)=F)$  using the recently-proposed method of quantile function regression \citep{si2021autoregressive}. 
%Defining this approach requires choosing a model class for the recalibrator $R$, a representation for its inputs $F=H(X)$, and a learning objective. 
%We define these below and refer to \citet{si2021autoregressive} for full details. 
The resulting method is shown in Algorithm \ref{alg:recal}.

\paragraph{Model}
We learn a model $R_\theta(\tau; \phi(F)) : \mathbb{R} \times \mathbb{R}^p \to \mathbb{R}$ of the inverse of the CDF of $\mathbb{P}(Y \mid H(X)=F)$.
%Our approach implements quantile function regression which fits a model of the inverse of the CDF.
Specifically, $R_\theta(\tau; \phi(F))$ takes in a scalar $\tau$ and features $\phi$, and outputs an estimate of the $\tau$-th quantile of $\mathbb{P}(Y \mid H(X)=F)$.
In our experiments, $R$ is parameterized by a fully-connected neural network with inputs $\tau$ and $\phi$.
%\vk{simplify?}

\paragraph{Features}

We form a $p$-dimensional representation  of $F$ by featurizing it via its quantiles $\phi(F) = (F^{-1}(\alpha_i))_{i=1}^p$ for some sequence of $p$ levels $\alpha_i$, typically uniform in $[0,1]$.
This parameterization works across diverse types of $F$, including parametric functions (e.g., Gaussians) or $F$ represented via a set of samples.
%The forecast $F$ can take multiple forms (e.g., Gaussians, sets of samples, etc.); we introduce a general parametrization $\phi(F)$ that works across diverse $F$. Specifically, 
%In the most general case, if we only have black-box access to a density function or a cumulative distribution function, we may form a $d$-dimensional representation by evaluating the distribution at a grid of $d$ points. For example, if we have black-box access to a quantile function $F^{-1}$, we may featurize $F$ via its sequence of quantiles $\phi(F) = (F^{-1}(\alpha_i))_{i=1}^d$ for some sequence of $d$ levels $\alpha_i$, possibly chosen uniformly in $[0,1]$.

\paragraph{Objective}

We train $R$ using the quantile proper loss $L_q$ % $L_q(F,y)=\Exp_{\tau\in U[0,1]}\rho_\tau(F,y)$ 
(see Table \ref{tbl:properlosses}).
% We experimented with various types of objectives $L$. Although any proper loss can be used, we found that the quantile proper loss (see Table \ref{tbl:properlosses}) was the most numerically stable and produced the best performance.
%
%We fit a model $R_\theta(\tau; \phi)$ to estimate the $\tau$-th conditional quantile $F^{-1}(\tau)$ at $\phi$. 
Specifically, when $R(\tau; \phi)$ is an estimate of the $\tau$-th quantile,
our objective becomes 
\begin{equation}
\Exp_{\phi, y} L_q(\phi, y) = \Exp_{\phi, y} \Exp_{\tau\in U[0,1]} \rho_\tau(R^{-1}(\tau; \phi), y),
\end{equation}
% (see \vk{link} for a definition of $\rho_\tau$). 
where $\rho_\tau$ is the check score (Equation \ref{eqn:checkscore}).
 We fit the objective using gradient descent, approximating the expectations using Monte Carlo. % (details are in the experiments section).
% The inner expectation corresponds to our data samples. For the outer expectation, we sample $\tau \sim U[0,1]$. We use a quantiles $\phi(F) = (F^{-1}(\alpha_i))_{i=1}^d$ to featurize $F$.%, like we did above.% [CITE OSTROVSKY ET AL].

\begin{algorithm}
  \caption{Distribution Calibrated Regression}
  \label{alg:recal}
  %\begin{algorithmic}
  \textbf{Input:}
    Pre-trained model $H : \mathcal{X} \to \Delta_\mathcal{Y}$, recalibrator $R : [0,1]  \times \Phi(\Delta_\mathcal{Y}) \to \mathbb{R}$, training set $\mathcal{D}$, calibration set $\mathcal{C}$\\
  \textbf{Output:}
    Recalibrated model $R \circ H : \mathcal{X} \to ([0,1] \to \mathbb{R})$ %\Delta_\mathcal{Y}$.
   \begin{enumerate}
%    \item Fit the base model on $\mathcal{D}$: $\min_{H} \sum_{(x, y) \in \mathcal{D}} L(H(x), y)$
    \item Create a training set for recalibrator: \\$\mathcal{S} = \{ (\phi(H(x)), y) \mid x, y \in \mathcal{C}\}$
    \item Fit the recalibrator $R$ on $\mathcal{S}$ using a quantile loss: \\
%    $\min_{R} \sum_{(\phi,y) \in \mathcal{S}} L_q\left(R(\phi), y\right)$ 
    $\min_{R} \sum_{(\phi,y) \in \mathcal{S}} \Exp_{\tau\in U[0,1]}\rho_\tau(R^{-1}(\tau;\phi),y)$
   \end{enumerate}
\end{algorithm}

%\subsection{Distribution Calibrated Classification via Density Estimation}
\subsection{Distribution Calibrated Classification}

We also propose an improved algorithm for the case when $Y$ is discrete. Our approach uses
the fact that the distribution of $Y$ can be represented via its probability mass function (PMF). 
We specify this algorithm below in terms of the model $R$, features $\phi$, and objective $L$. 
%When $Y$ is discrete, the distribution $\mathbb{P}(Y \mid H(X)=F)$ is represented via its probability mass function (PMF). Learning a model $R$ of the PMF again involves specifying parametric forms for $R$ and $F$, as well as a learning objective. 

\paragraph{Features, Model, and Objective}
In classification, each $F \in \Delta_\mathcal{Y}$ is categorical and can be represented as a vector $p_F \in \Delta_K$ of $K \geq 2$ class membership probabilities living in a simplex $\Delta_K$ over $K$-dimensional probability vectors.

%\paragraph{Recalibration Model}

A natural architecture for $R$ is a sequence of dense layers mapping the simplex $\Delta_K$ into $\Delta_K$.
Such network recalibrators can be implemented easily within deep learning frameworks and work well in practice despite being non-convex, as we later demonstrate empirically.

%\paragraph{Learning Objective}
A natural choice of proper scoring rule is the log-loss $L_\text{log}$. Optimizing it is a standard supervised learning problem. 

\begin{algorithm}
  \caption{Distribution Calibrated Classification}
  \label{alg:recal2}
  %\begin{algorithmic}
  \textbf{Input:}
    Pre-trained model $H : \mathcal{X} \to \Delta_K$, recalibrator $R : \Delta_K \to \Delta_K$, training set $\mathcal{D}$, calibration set $\mathcal{C}$\\
  \textbf{Output:}
    Recalibrated model $R \circ H : \mathcal{X} \to \Delta_K$.
   \begin{enumerate}
%    \item Fit the base model on $\mathcal{D}$: $\min_{H} \sum_{(x, y) \in \mathcal{D}} L(H(x), y)$
    \item Create a recalibrator training set: \\$\mathcal{S} = \{ (p_{H(x)}, y) \mid x, y \in \mathcal{C}\}$
    \item Fit the recalibration model $R$ on $\mathcal{S}$: \\
    $\min_{R} \sum_{(p,y) \in \mathcal{S}} L_\text{log}\left(R(p), y\right)$ 
   \end{enumerate}
\end{algorithm}

\subsection{Practical Considerations}

\paragraph{Diagnostic Tools}

We propose assessing calibration in a regression setting using the check score $\rho_\tau$ aggregated over $q$ quantiles $\{\tau_i\}_{j=1}^q$ (typically uniform in $[0,1]$) and a dataset $\mathcal{D}=\{x_i, y_i\}_{i=1}^n$ as 
$\frac{1}{n}\sum_{i=1}^n \frac{1}{q}\sum_{j=1}^q \rho_{\tau_j}(F_{x_i}, y_i)$,
where $F_{x_i}$ is the forecast at $x_i$.
Each $\rho_\tau$ is a consistent estimator for $\tau$-th quantile---small $\rho_\tau$ indicate that the target $y$ falls below the $\tau$-th quantile a fraction $\tau$ of the time.

We recommend assessing the overall performance of $F$ via a proper loss---this measures both calibration and sharpness and oftentimes corresponds to standard notions of accuracy (e.g., the log-loss often reduces to the mean squared error). Sharpness can be assessed heuristically via the variance of $F$, although mathematically, this normally does not correspond to a component of a proper loss.

\paragraph{Calibrating Bayesian Models}

When the model $H$ is Bayesian or consists of an ensemble, the forecasts $F$ may not have a closed form expression---instead, we may only sample from $F$. In such cases, we may compute features $\phi$ on samples from $F$---e.g. by computing empirical quantiles.

\paragraph{Non-Parametric Recalibrators}
Algorithms \ref{alg:recal0} represents a general framework that admits a range of recalibration models and objectives.
One possible additional class of model families for $R$ are non-parametric kernel density estimators over a low-dimensional space induced by features $\phi(F)$.
%Ideal recalibrators are highly effective at optimizing the proper loss $L$ (see Section \ref{sec:recalibration}).
%In a simple setting like binary classification, our task reduces to one-dimensional density estimation; in such cases we can provably achieve calibration asymptotically by using kernel density estimation for the recalibrator $R$, while controlling the kernel width as a function of the dataset size to trade off overfitting and underfitting \citep{wasserman2006all}. In regression settings, we may rely on other non-parametric techniques such as Gaussian processes.
%
Such non-parametric models can be combined with parameterized neural networks via architectures such as mixture density networks (MDN; \citet{bishop1994mixture}).

%% file: recalibration.tex
\section{Theoretical Analysis}\label{sec:theory}

%% Predictive uncertainties in machine learning should maximize sharpness subject to being calibrated \citep{gneiting2007probabilistic}.
%Next, we now show that under some assumptions, calibrated and sharp uncertainties can be obtained in  modern machine learning models
%via our simple procedure.
%% in a black-box manner and without sacrificing overall performance.
%% methods based on deep learning.
%% In particular, it can be done in a {\em black-box} manner over any existing model without sacrificing performance in other ways.
%In that sense, distribution calibration may be easier to obtain than was previously thought.

Next, we formally establish the correctness of our procedure. Our main claim is that constructing a distribution calibrated model is equivalent to estimating the density $\mathbb{P}(Y = y \mid F_X = F)$. 
When $F$ is represented via low-dimensional features $\phi$, this problem is tractable---hence calibration may be easier to enforce than previously thought.
%This density estimation problem will typically be low-dimensional since the distribution is conditioned on $F$, which in practice admits a low-dimensional parameterization (e.g., $\mu, \sigma^2$ for a Gaussian).
%Low-dimensional density estimation is a well-studied problem with mature and provably correct algorithms; in that sense, distribution calibration may be easier to maintain than was previously thought.

% separate this out into a proof section and an interpretation (calibration is a free lunch) section?
% \subsection{Calibration is a Free Lunch}

% Next, we prove that Algorithm \ref{alg:recal} indeed produces models that perform calibrated risk minimization. 

% We start with some notation. 
% Let $L$ be a proper loss and let $L_c$ be its associated calibration loss derived from the calibration-reliability decomposition of $L$. 
% Let $\mathcal{L}^{(H)}, \mathcal{L}^{(H)}_c$ denote the expected values of $L, L_c$ for a classifier $H$:
% \begin{align*}
%     \mathcal{L}^{(H)} = \mathbb{E}_{X,Y}[L(H(X),Y]
%     & & \mathcal{L}_C^{(H)} = \mathbb{E}_{X,Y}[L_C(H(X),Q],
% \end{align*}
% where $Q(y) := \mathbb{P}(Y=y \mid H(X))$ is the conditional distribution of $Y$ given a forecast of $H(X)$. Let $\mathcal{L}^*$ denote the optimal value of $\mathcal{L}^{(H)}$.
\paragraph{Notation}
We start with some notation. 
We have a calibration dataset $\mathcal{C}$ of size $m$ sampled from $\mathbb{P}$ and train a recalibrator $R : \Delta_\mathcal{Y} \to \Delta_\mathcal{Y}$
over the outputs of a base model $H$ to minimize a proper loss $L$. We denote the Bayes-optimal recalibrator by $B := \mathbb{P}(Y=y\mid H(X))$;
the distribution of $Y$ conditioned on the forecast $(R \circ H)(X)$ is $Q := \mathbb{P}(Y=y \mid (R \circ H)(X))$.
We are interested in expectations of 
various losses over $X,Y$; to simplify notation, we omit the variable $X$, e.g. $\mathbb{E}[L(R \circ H,Y)] = \mathbb{E}[L(R(H(X)),Y)]$.
 
\subsection{Distribution Calibration}
 
Our first claim is that distribution calibration reduces to a density estimation problem.
%if we can perform density estimation, we can perform distribution calibration.
We first formally define the task of density estimation.
\begin{task}[Density Estimation]
\label{ass:density}
The model $R$ approximates the distribution $B := \mathbb{P}(Y=y\mid H(X))$ if the proper loss of $R$ tends to that of $B$ as $m \to \infty$ and w.h.p. for all $m$:
% The recalibrator $R$ learns a Bayes-optimal density by minimizing the proper loss $L$. 
% Given a dataset of size $T$, and a proper loss $L$ the recalibrator $R$ learns to approximate 
% The Bayes-optimal recalibrator $B(Y=y\mid H(X))$ such that w.h.p.~we have
\begin{align*}
\mathbb{E}[L(B\circ H,Y)] 
\leq \mathbb{E}[L(R \circ H,Y)] < \mathbb{E}[L(B \circ H,Y)] + \delta_m
    % L^* \leq \mathbb{E}[L((R\circ H)(X),Y)] \leq L^* + \delta
\end{align*}
% \mathbb{E}[L(R(H(X)),Y)]
where $\delta_m> 0$, $\delta_m = o(m)$ is a bound decreasing with $m$.% and $\mathbb{E}[L(B\circ H,Y)]$ is the irreducible loss.
\end{task}

When the features $\phi(F)$ are low-dimensional (e.g., $F$ is Gaussian), this task can be solved efficiently.
%This assumption implies that the recalibrator can perform density estimation  in what is usually a small number of dimensions (one or two). 
We show via experiments that parametric neural network density estimators are effective at Task \ref{ass:density}, and their performance can be verified on a held-out dataset.
%Although parametric neural network density estimators are not necessarily guaranteed to solve this task (due to non-convexity), we demonstrate empirically that in practice they do, and also one can evaluate their performance on a held out dataset.
Alternatively, non-parametric kernel density estimation is formally guaranteed to estimate this density given enough data \citep{wasserman2006all}.
%For some recalibrators, e.g., neural nets, it may not provably hold (e.g., because of non-convexity). However, neural networks are effective density estimators in practice, and we can quantify whether they estimate density well on a hold-out set.
%This assumption provably holds for many non-parametric density estimation methods.

%\begin{fact}[\citet{wasserman2006all}]
%When $R$ is a kernel density estimation algorithm \vk{name} and $L$ is the log-loss, Task \ref{ass:density} is solved with $\delta=o(1/m^{2/3})$. \vk{remove?}
%\end{fact}
% such as kernel density estimation \citep{wasserman2006all}. It may even hold when $R$ is a sufficiently expressive neural network, although we cannot prove it.
% Note that rate of $\delta$ is good when dimensionality is low.

%We now prove two key lemmas. We show that Algorithm \ref{alg:recal} outputs calibrated forecasts without reducing the performance of the base model, as measured by regret relative to loss $L$.

Our next proposition states that minimizing the recalibration objective defined in Algorithm \ref{alg:recal0} yields a model $R$ that performs density estimation (Task \ref{ass:density}).

\begin{proposition}%[Calibration]
\label{lem:density}
Suppose that $R$ minimizes the objective in Step 2 of Algorithm \ref{alg:recal0}. Then
$R$ estimates the density (\ref{eqn:density}) and performs Task \ref{ass:density}.
\end{proposition}
\begin{proof}[Proof (Sketch)]
%We learn this density by optimizing a proper scoring rule $S(\cdot, \cdot)$. We give examples of specific scoring rules below. 
Algorithm \ref{alg:recal0} outputs a recalibrator $R$ that minimizes the objective $\sum_{x, y \in \mathcal{C}} L(R(F_x), y)$ over a calibration dataset $\mathcal{C}$ sampled i.i.d.~from $\mathbb{P}$. Observe that
\begin{align}
\sum_{x, y \in \mathcal{C}} L(R(F_x), y) 
& \approx \Exp_{F_X \sim \mathbb{P}} \Exp_{Y \sim \mathbb{P}(Y|F_X)} [ L(R(F_X), Y) ] \nonumber\\
& = \Exp_{F_X \sim \mathbb{P}} [ L(R(F_X), \mathbb{P}(Y|F_X)) ]. \label{eqn:expected_loss}
\end{align}
%where $F_X = H(X)$. 
%The first line follows by Monte Carlo; the second holds because $L$ is proper. It follows from empirical risk minimization and the properties of Monte Carlo estimation that minimizing the empirical loss $\sum_{x, y \in \mathcal{C}} L(R(F_x), y)$ yields a minimizer of (\ref{eqn:expected_loss}), which is $\mathbb{P}(Y|F_X))$ since $L$ is proper.
where $F_X = H(X)$. The first line follows by Monte Carlo and the second line follows from the definition of a proper loss. It follows from empirical risk minimization and the properties of Monte Carlo estimation that minimizing the empirical loss $\sum_{x, y \in \mathcal{C}} L(R(F_x), y)$ yields a minimizer of (\ref{eqn:expected_loss}), which is $\mathbb{P}(Y|F_X))$ since $L$ is a proper loss.
\end{proof}

Next, we prove that using an $R$ that estimates density (\ref{eqn:density}) (Task \ref{ass:density}) yields models that are asymptotically calibrated, i.e. their calibration error tends to zero as $m \to \infty$.

\begin{proposition}%[Calibration]
\label{lem:calibration}
Let $R$ be a model solving Task \ref{ass:density}.
Then $R \circ H$ is asymptotically calibrated and there exists $\delta_m = o(m)$ such that
$\mathbb{E}[L_c(R \circ H,Q)] < \delta_m$ for all $m$  w.h.p.
% $\mathcal{L}_C \leq \epsilon$ when $T \to \infty$.
\end{proposition}
\begin{proof}
Recall that the loss $\mathbb{E}[L(R \circ H,Y)]$ decomposes into a sum of calibration and refinement terms $\mathbb{E}[L_c(R \circ H,Q)] + \mathbb{E}[L_r(Q)]$ where $Q(y) := \mathbb{P}(Y=y \mid (R \circ H)(X))$.

As shown by \citet{kull2015novel}, refinement further decomposes into a group loss and an irreducible term:
$\mathbb{E}[L_r(Q)] = \mathbb{E}[L_g(Q,B\circ H)] + \mathbb{E}[L(B\circ H,Y)],$
where $B(Y=y\mid H(X))$ is the Bayes-optimal recalibrator. The form of the group loss $L_g$ is the same as that of $L_c$.
We may then write:
\begin{align*}
& \underbracket{\mathbb{E}[L(B\circ H,Y)]}_\text{Bayes-Optimal Loss} \\
& \leq \underbracket{\mathbb{E}[L_c(R \circ H,Q)]}_\text{Calibration Loss} + \underbracket{\mathbb{E}[L_g(Q,B\circ H)]}_\text{Group Loss} + \underbracket{\mathbb{E}[L(B\circ H,Y)]}_\text{Bayes-Opt Loss} \\
& = \underbracket{\mathbb{E}[L(R \circ H,Y)]}_\text{Proper Loss} 
< \underbracket{\mathbb{E}[L(B \circ H,Y)]}_\text{Bayes-Optimal Loss} + \delta
    % L^* \leq \mathbb{E}[L((R\circ H)(X),Y)] \leq L^* + \delta
\end{align*}
where $\delta_m>0, \delta_m=o(m)$. 
In the first equality we used the decomposition of \citet{kull2015novel} and in the last inequality we used Assumption \ref{ass:density}.
It follows that $\mathbb{E}[L_c(R \circ H,Q)] < \delta_m$, i.e. the calibration loss is small.
%
% Let $L, L_c, L_g, L^*$ respectively denote the expected values of the original loss, the calibration term, the group term, and the irreducible term. For a good $R$, by Assumption \ref{ass:density}, we have $L < L^* + \delta$ w.h.p. Since
% $$L^* \leq L_c + L_g + L^* = L  \leq L^* + \delta,$$
% we must have that $L_c \leq \delta$.
\end{proof}

\paragraph{Practical Implications.}

% We highlight a number of design choices informed by our theoretical analysis. 

Assumption \ref{ass:density} suggests that we want to use a model family that can minimize the expected risk $\mathbb{E}[L(H(X),Y)]$ well. 
Thus, in practice we want to select highly flexible algorithms for which we can control overfitting and underfitting. This motivates our earlier advice of using density estimation algorithms---which have provable guarantees---and neural networks---which are expressive and feature effective regularization techniques. 

% In a simple setting like binary classification, this corresponds to one-dimensional density estimation; in such cases we can provably achieve calibration asymptotically by using kernel density estimation for the recalibrator $R$, while controlling the kernel width as a function of the dataset size to trade off overfitting and underfitting \citep{wasserman2006all}.

% When $H$ outputs arbitrary distributions, we still want to minimize the expected risk $\mathbb{E}[L(H(X),Y)]$. This motivates the use of a model family for $R$ that is sufficiently expressive on large datasets, and can be regularized on smaller datasets. Furthermore, we want the optimization over the set of $R$ to be tractable and easy to implement in practice.

\subsection{Vanishing Regret}

Our second result shows that our approach yields calibration without degrading the performance of the baseline model. We define performance using a proper loss $L$; for example, if $L$ is the log-likelihood we cover most standard performance metrics such as the L2 or the cross-entropy losses.

\begin{proposition}
\label{lem:loss}
The recalibrated model has asymptotivally vanishing regret relative to the base model: $\mathbb{E}[L(R \circ H,Y)] \leq \mathbb{E}[L(H,Y)] + \delta_m,$ where $\delta_m >0, \delta=o(m)$ is a bound that decreases with $m$.
\end{proposition}

\begin{proof}
%The claim holds by empirical risk minimization. Since $R \circ H$ minimizes $L$, but is more expressive than $H$ and $R$ can represent the identity map (by Assumption \ref{ass:density}).
If we solve Task \ref{ass:density}, then $\mathbb{E}[L(R \circ H,Y)] \leq \mathbb{E}[L(B \circ H,Y)] + \delta_m \leq \mathbb{E}[L(H,Y)] + \delta_m$, where the second inequality holds because the Bayes-optimal recalibrator $B$ achieves an equal or lower loss than an identity mapping.
\end{proof}

%We now combine these two lemmas to show Algorithm \ref{alg:recal} ensures calibration and low regret.
%
%\begin{theorem}
%\label{thm:recal}
%Algorithm \ref{alg:recal} produces a model that minimizes expected risk, while w.h.p. achieving asymptotically optimal calibration.
%% The calibrated risk minimization principle is achieved by Algorithm \ref{alg:recal} w.h.p. as $T \to \infty$.
%\end{theorem}
%\begin{proof}
%The base model $H$ is trained using empirical risk minimization (ERM). The model $R \circ H$ minimizes the same objective $L$, hence minimizes the same expected risk by ERM theory.
%% hence and 
%Also, by Lemma \ref{lem:loss}, the expected risk of $R \circ H$ also asymptotically approaches a lower value as that of $H$.
%
%By Lemma \ref{lem:calibration}, the model $R \circ H$ produces asymptotically calibrated forecasts w.h.p.
%\end{proof}

% Let's emphasize again the significance of Theorem \ref{thm:recal}. We can take any base model and apply Algorithm \ref{alg:recal} with a number of off-the-shelf density estimation methods as the model $R$. 
Thus, given enough data, we are guaranteed to produce calibrated forecasts and preserve base model performance (as measured by $L$).
Note that there is no guarantee that sharpness will not worsen (it likely will)---our claims only pertain to the proper loss.
% in fact, we expect it to worsen most of the time. 
%Our work only claims that the {\em proper loss} of the original model will be approximately unchanged
% Thus, calibration is a property that can be achieved in most applications of machine learning with almost no cost. As such, calibration is a rare free lunch in machine learning.

\paragraph{Finite-Sample Bounds.}

Note that our analysis provides {\em finite-sample} and not only asymptotic bounds on the regret and calibration error---the bounds are stated in terms of $\delta$, which is $o(m)$. The bound $\delta$ on the calibration error directly depends on the finite-sample bound on the generalization error of the algorithm used as the recalibrator.

%% file: calibration.tex
%\section{What Uncertainties Are Needed in Modern Deep Learning?}\label{sec:calibration}
\section{On Calibration in Machine Learning}\label{sec:calibration}

In this section, we discuss the role of calibration in machine learning.
%In this section, we argue for why our calibrated training procedures have potential to benefit machine learning systems.
%In this section, we discuss ways in which our calibrated training procedures can benefit machine learning systems.
We argue that calibration may be simpler to obtain than previously thought and enforcing it in predictive models unlocks benefits across downstream applications.

% \paragraph{Training Calibrated Models.} 
% \subsection{Do Machine Learning Models Produce Calibrated And Sharp Forecasts?}
\subsection{Are Models Calibrated Out-of-the-Box?}
%\vk{Put this in section on calibraiton}
Machine learning models typically make uncalibrated probabilistic predictions \citep{niculescu2005predicting,guo2017calibration}.
One reason for this is the limited expressivity of the model $H$---we cannot assign the correct probability to every level curve of the distribution. Another reason is the use of computational approximations (e.g., variational inference) to compute intractable predictive uncertainties.
%Recalibration methods train an auxiliary model $R : [0,1] \to [0,1]$ on top of a pre-trained forecaster $H$ such that $R \circ H$ is calibrated.
%

While most models $H$ are typically trained with a proper loss (the log-likelihood),
this rarely yields calibration.
The proper loss equals a sum of 
calibration and sharpness terms---not being able to perfectly optimize the loss, the model strikes a balance between suboptimal calibration and sharpness.
Our method instead estimates densities in low dimensions (in the space of outputs of $H$), which tractably approaches a Bayes-optimal loss and yields predictions with low calibration error.

%A final reason stems from how models are trained---since we cannot fit a perfect $H$, standard objective functions induce a tradeoff between sharp and calibrated forecasts. 
%We show how to work around this issue via specialized training procedures.

\subsection{On the Simplicity of Ensuring Calibration}

Machine learning models are normally trained  to minimize expected risk; 
we argue that in addition they should be calibrated. 
Calibration is achievable using the following procedure:
(1) train a base model $H$ on the main dataset $\mathcal{D}$;
(2) train a recalibrator $R$ using Algorithms \ref{alg:recal} or \ref{alg:recal2}.

%While models are typically not calibrated out-of-the-box, 
Our results show that calibration can be provably achieved if we can perform density estimation in low dimensions (Task \ref{ass:density}). Furthermore, the recalibrated forecasts do not degrade performance, as measured by a proper loss. In that sense, incorporating calibrated forecasting in machine learning systems may be easier than previously thought.

Our methods can be seen as an implementation of the principle of
\citet{gneiting2007probabilistic}, who argue that 
predictive uncertainties should be maximally sharp subject to being calibrated.
While this approach is used in statistics for {\em evaluating} probabilistic models, our methods provides a way of enforcing this principle.
\subsection{On the Importance of Calibrated Predictinos}\label{sec:discussion}

% \subsection{Why Do We Need Calibrated And Sharp Uncertainties?}

Probabilistic models
are key building blocks of machine learning systems
in many areas---medicine, robotics, industrial automation, and more.
%Calibration is not difficult to achieve in many of these domains; 
We argue that maintaining calibration in predictive models will unlock benefits in downstream applications across may of these domains.

\paragraph{Safety and Interpretability.}
Good predictive uncertainties are important for model interpretability: in user-facing applications, humans make decisions based on model outputs and need to assess the confidence of the model, for example when interpreting an automated medical diagnosis. Calibration is also important for model safety: in areas such as robotics, we want to minimize the probability of adverse outcomes (e.g., a crash), and estimating their probability is a key step towards averting them \citep{Berkenkamp2017}.

\paragraph{Model-Based Planning.}
More generally, good predictive uncertainties also improve downstream decision-making applications such as model-based planning \citep{malik2019calibrated}, a 
setting in which agents learn a model of the world to plan future decisions \citep{deisenroth2011pilco}.
Planning with a probabilistic model improves performance and sample complexity, especially when representing the model using a deep neural network.
and improves the cumulative reward and the sample complexity of model-based agents~\citep{rajeswaran2016epopt,chua2018}.

\paragraph{Efficient Exploration.}
Balancing exploration and exploitation is a central challenge in reinforcement learning, Bayesian optimization, and active learning. 
When probabilistic models are uncalibrated, inaccurate confidence intervals might incentivize the model to explore ineffective actions, degrading performance.
Calibrated uncertainties have been shown to improve decision-making in bandits \citep{malik2019calibrated}, Bayesian optimization \citep{deshpande2021calibration}, and are likely to extend to active learning.

%% file: experiments.tex
\begin{table*}[htb]
\vspace{-1mm}
\begin{center}
\scriptsize
\begin{tabular}{l|rr|rr|rr|rr}
\toprule
{} & \multicolumn{8}{l}{Bayesian Linear Regression} \\
{} & \multicolumn{2}{l}{Uncalibrated} & \multicolumn{2}{l}{Kuleshov et al.} & \multicolumn{2}{l}{Song et al.} & \multicolumn{2}{l}{Ours} \\
{} &                        MAE &       CHK &             MAE &       CHK &         MAE &       CHK &     MAE &       CHK \\
dataset &                            &                 &                 &                &             &                 &                 &         \\
\midrule
mpg     &                      2.45 $\pm$ .00 &    0.92 $\pm$ .00 &           2.46 $\pm$ .00 &   0.91 $\pm$ .00 &       2.49 $\pm$ .01 &    {\bf 0.91 $\pm$ .00} &   2.39 $\pm$ .03 &    {\bf 0.90 $\pm$ .01} \\
boston  &                      3.45 $\pm$ .00 &    1.39 $\pm$ .00 &           3.39 $\pm$ .00 &  1.36 $\pm$ .00 &       3.38 $\pm$ .02 &   1.37 $\pm$ .01 &   3.34 $\pm$ .03 &    {\bf 1.31 $\pm$ .02} \\
yacht   &                      6.17 $\pm$ .00 &    2.43 $\pm$ .00 &           5.96 $\pm$ .00 &   2.37 $\pm$ .00 &       1.50 $\pm$ .01 &    1.27 $\pm$ .01 &   0.90 $\pm$ .01 &    {\bf 0.35 $\pm$ .01} \\
wine    &                      0.62 $\pm$ .00 &    0.24 $\pm$ .00 &           0.62 $\pm$ .00 &   0.23 $\pm$ .00 &       0.63 $\pm$ .00 &    {\bf 0.24 $\pm$ .00} &   0.62 $\pm$ .01 &    {\bf 0.24 $\pm$ .01} \\
crime   &                      0.52 $\pm$ .00 &    0.20 $\pm$ .00 &           0.51 $\pm$ .00 &   0.20 $\pm$ .00 &       0.52 $\pm$ .00 &    0.20 $\pm$ .00 &   0.51 $\pm$ .01 &    {\bf 0.19 $\pm$ .00} \\
auto    &                      0.64 $\pm$ .00 &    0.25 $\pm$ .00 &           0.62 $\pm$ .00 &    0.25 $\pm$ .00 &       0.63 $\pm$ .00 &    {\bf 0.25 $\pm$ .00} &   0.63 $\pm$ .01 &    {\bf 0.25 $\pm$ .01} \\
cpu     &                     39.9 $\pm$ 0.00 &   15.8 $\pm$ 0.00 &          39.0 $\pm$ 0.00 &  15.4 $\pm$ 0.00 &      35.1 $\pm$ 0.24 &  {\bf 13.9 $\pm$ 0.05} &  28.1 $\pm$ 0.58 &   {\bf 13.6 $\pm$ 0.45} \\
bank    &                     39.5 $\pm$ 0.00 &   17.4 $\pm$ 0.00 &          39.1 $\pm$ 0.00 & 16.6 $\pm$ 0.00 &      34.1 $\pm$ 0.19 &  15.4 $\pm$ 0.04 &  29.3 $\pm$ 0.61 &  {\bf 14.2 $\pm$ 0.42} \\
\bottomrule
\end{tabular}
\begin{tabular}{l|rr|rr|rr|rr}
\toprule
{} & \multicolumn{8}{l}{Bayesian Neural Network} \\
{} & \multicolumn{2}{l}{Uncalibrated} & \multicolumn{2}{l}{Kuleshov et al.} & \multicolumn{2}{l}{Song et al.} & \multicolumn{2}{l}{Ours} \\
{} &                     MAE &       CHK &             MAE &       CHK &         MAE &       CHK &     MAE &        CHK \\
dataset &                         &                &                 &                 &             &                 &                 &         \\
\midrule
mpg     &                   2.73 $\pm$ .11 &   1.19 $\pm$ .07 &           2.97 $\pm$ .13 &    1.17 $\pm$ .07 &       2.67 $\pm$ .15 &    {\bf 1.10 $\pm$ .08} &   2.60 $\pm$ .18 &   {\bf 1.08 $\pm$ .08} \\
boston  &                   2.97 $\pm$ .15 &   1.23 $\pm$ .07 &           3.00 $\pm$ .16 &    1.20 $\pm$ .08 &       3.30 $\pm$ .18 &   {\bf 1.18 $\pm$ .09} &   2.98 $\pm$ .17 &    {\bf 1.19 $\pm$ .10} \\
yacht   &                   3.54 $\pm$ .19 &    1.53 $\pm$ .06 &           3.77 $\pm$ .22 &    1.51 $\pm$ .07 &       3.37 $\pm$ .25 &   { 1.51 $\pm$ .07} &   3.17 $\pm$ .21 &    {\bf 1.41 $\pm$ .07} \\
wine    &                   0.62 $\pm$ .03 &    0.25 $\pm$ .01 &           0.63 $\pm$ .03 &    0.24 $\pm$ .02 &       0.62 $\pm$ .04 &   {\bf 0.23 $\pm$ .01} &   0.62 $\pm$ .04 &    {\bf 0.23 $\pm$ .02} \\
crime   &                   0.50 $\pm$ .01 &    0.19 $\pm$ .00 &           0.48 $\pm$ .01 &    0.19 $\pm$ .01 &       0.48 $\pm$ .01 &  0.19 $\pm$ .01 &   0.49 $\pm$ .01 &    {\bf 0.18 $\pm$ .00} \\
auto    &                   0.62 $\pm$ .02 &    0.25 $\pm$ .01 &           0.62 $\pm$ .02 &    0.24 $\pm$ .01 &       0.64 $\pm$ .02 &    0.24 $\pm$ .01 &   0.64 $\pm$ .02 &    {\bf 0.23 $\pm$ .00} \\
cpu     &                  74.0 $\pm$ 1.85 &   35.5 $\pm$ .92 &          71.0 $\pm$ 1.83 &  28.6 $\pm$ .93 &      68.4 $\pm$ 1.85 &   {\bf 31.8 $\pm$ 1.22} &  66.4 $\pm$ 1.91 &   {\bf 31.3 $\pm$ 1.10} \\
bank    &                  96.0 $\pm$ 1.89 &   46.0 $\pm$ .94 &          90.8 $\pm$ 1.87 &     44.1 $\pm$ .95 &      87.0 $\pm$ 1.88 &  41.1 $\pm$ 1.18 &  85.0 $\pm$ 1.90 &  {\bf 39.2 $\pm$ 1.09} \\
\bottomrule
\end{tabular}
\begin{tabular}{l|rr|rr|rr|rr}
\toprule
{} & \multicolumn{8}{l}{Deep Ensemble} \\
{} & \multicolumn{2}{l}{Uncalibrated} & \multicolumn{2}{l}{Kuleshov et al.} & \multicolumn{2}{l}{Song et al.} & \multicolumn{2}{l}{Ours} \\
{} &           MAE &       $\;\;\;$CHK &             MAE &       $\;\;\;$CHK &         MAE &        $\;\;\;$CHK &      MAE &         $\;\;\;$CHK \\
dataset &               &                 &                 &                 &             &                  &                  &          \\
\midrule
mpg     &         7.66 $\pm$ .32 &   3.55 $\pm$ .12 &          11.8 $\pm$ .41 &   5.20 $\pm$ .21 &       9.57 $\pm$ .39 &   3.93 $\pm$ .23 &    9.01 $\pm$ .40 &   {\bf 3.53 $\pm$ .18} \\
boston  &         8.42 $\pm$ .31 &  3.82 $\pm$ .18 &          12.8 $\pm$ .52 &   5.49 $\pm$ .27 &       8.13 $\pm$ .44 &   3.91 $\pm$ .21 &    8.00 $\pm$ .46 &   {\bf 3.62  $\pm$ .20}\\
yacht   &         9.40 $\pm$ .38 &   4.60 $\pm$ .11 &          10.8 $\pm$ .49 &   4.93 $\pm$ .23 &       9.45 $\pm$ .42 &   4.41 $\pm$ .15 &    9.36 $\pm$ .51 &   {\bf 4.27 $\pm$ .15}\\
wine    &         0.71 $\pm$ .02 &   0.30 $\pm$ .01 &           0.70 $\pm$ .03 &   0.27 $\pm$ .01 &       0.69 $\pm$ .03 &    {\bf 0.26 $\pm$ .01} &    0.69 $\pm$ .03 &   {\bf 0.26 $\pm$ .01} \\
crime   &         0.68 $\pm$ .02 &   0.25 $\pm$ .01 &           0.77 $\pm$ .03 &    0.34 $\pm$ .02 &       0.68 $\pm$ .02 &  {\bf 0.25 $\pm$ .01} &    0.68 $\pm$ .03 &   {\bf 0.25 $\pm$ .01} \\
auto    &         0.86  $\pm$ .02 &    0.30 $\pm$ .01 &           0.86 $\pm$ .03 &   0.31 $\pm$ .01 &       0.87 $\pm$ .03 &    0.35 $\pm$ .02 &    0.86 $\pm$ .04 &    {\bf 0.30 $\pm$ .02} \\
cpu     &     58.1 $\pm$ 1.77 &  20.0 $\pm$ 1.01 &          57.3 $\pm$ 1.82 &  18.3 $\pm$ 1.00 &      57.9 $\pm$ 1.81 &  {\bf 18.3 $\pm$ 0.94} &  55.4 $\pm$ 1.89 &  {\bf 17.4  $\pm$ 1.08}\\
bank    &    59.0 $\pm$ 1.81 &  21.2 $\pm$ 0.99 &          58.7 $\pm$ 1.85 &   21.0 $\pm$ 0.98 &      57.4 $\pm$ 1.87 & {\bf 20.0  $\pm$ 0.96} &  55.8 $\pm$ 1.92 &  {\bf 19.1  $\pm$ 1.02} \\
\bottomrule
\end{tabular}
\end{center}
\caption{\footnotesize Calibration and accuracy on UCI regression datasets. We evaluate Bayesian linear regression, Bayesian neural networks, and deep ensembles using the mean average error (MAE) and check score (CHK), comparing against \citet{kuleshov2018accurate} and \citet{song2019distribution}. These metrics precisely capture the two properties studied in our paper: the ability of our methods to ensure good calibration while having a very small regret relative to the baseline model as measured by a supervised learning loss.}\label{tbl:uci}
% \vspace{-10mm}
\end{table*}

\section{Experiments}

\subsection{Setup}

\paragraph{Datasets.} 

%We validate our approach on several machine learning datasets.
%We validate our approach on eight UCI regression datasets, several of which were used in \cite{lakshminarayanan2016simple}. 
We use a number of UCI regression datasets varying in size from 194 to 8192 training instances; each training input may have between 6 and 159 continuous features. We randomly use 25\% of each dataset for testing, and use the rest for training. 
% We report averages over 5 random splits. 
We also perform classification on the following standard datasets: MNIST, SVHN, CIFAR10. 
% estimation on the larger Make3D dataset \cite{saxena2009make3d}, using the setup of \citet{kendall2017uncertainties}.

% %In addition, we test our method on a time series forecasting task. 
% We also test our method on time series forecasting and reinforcement learning tasks.
% We use daily grocery sales from the Corporacion Favorita Kaggle dataset; we forecast the highest-selling item (\#1503844) and use data from 2014-01-01 to 2016-05-31 in stores \#1-4 for training and data from 2016-06-01 to 2016-12-31 for testing. We use auto-regressive features from the past four days as well as binary indicators for the day of the week and the week of the year.
% %On all datasets, continuous features were normalized to have zero mean and unit variance.

\paragraph{Models.}

Our first model is Bayesian Ridge Regression \citep{mackay1992bayesian}. It uses a spherical Gaussian prior over the weights and a Gamma prior over the precision parameter. Posterior inference is performed in closed form as the prior is conjugate.

We also test a number of deep neural networks. We use variational dropout \citep{Gal2016Dropout} to produce probabilistic predictions. In our UCI experiments, we use fully-connected feedforward neural networks with two layers of 128 hidden units with a dropout rate of 0.5 and parametric ReLU non-linearities. We use convolutional neural networks (CNNs) on the image classification tasks. These are formed by fine-tuning a ResNet50 architecture on the training split for each dataset.

% In our recalibration experiments, we fit a base model on the training set, and then use a second smaller neural network to recalibrate on the same training set. We didn't use a separate recalibration set as we did not observe overfitting. 

We additionally compare against a popular uncertainty estimation method recently developed specifically for deep learning models: deep ensembles \citep{lakshminarayanan2016simple}. Deep ensembles average the predictive distributions of multiple models; we ensembled 5 neural networks, each having the same architecture as our standard model.

% \paragraph{Recalibrator.}

Our recalibrator $R$ was also a densely connected neural network with two fully connected hidden layers of 20 units each and parametric ReLU non-linearities. We added dense skip connections between the layers. 
We held out 15\% of the training set (up to max of 500 datapoints) for recalibration. 
% This leaves less data for training, but 
On large datasets 500 points, is a relatively small quantity. When data is scarce, we can fit an ensemble of models calibrated on leave-one-out folds.

In regression experiments, we featurized input distributions $F$ using nine quantiles $[0.1,...,0.9]$. We trained $R$ using the quantile regression objective of Algorithm \ref{alg:recal}; we concatenated the quantile parameter $\tau \in [0,1]$ to the featurization of $F$.
In classification experiments, the inputs and the ouputs of $R$ are class probabilities, and $R$ is trained using log-likelihood maximization defined in Algorithm \ref{alg:recal2}. All other architectural details are unchanged.
% implemented we used a mixture density network with up $K\leq3$ mixture components in the final layer. 
% We trained the recalibrator 10 times on each experiment for each value of $1\leq K \leq 3$ and reported the performance of the model with the best training performance. This helped address numerical instabilities during training.
% Training was unstable across all the datasets, hence we reran the recalibrator 5 times on each experiment and reported the performance of the model with the best training performance.

We did not observe significant overfitting in our experiments. We believe overfitting is mitigated by the fact that we perform quantile regression and thus learn a complex distribution function that is not easy to overfit.

% \paragraph{Baselines.}

% We use as baselines two popular uncertainty estimation methods recently developed specifically for deep learning models: concrete dropout \citep{gal2017concrete} and deep ensembles \citep{lakshminarayanan2016simple}. Concrete dropout is a more modern variant of variational dropout in which dropout probabilities are learned from data; we replace standard dropout in our models with concrete dropout. Discrete ensembles average the predictive distributions of multiple models; we ensembled 5 neural networks, each having the same architecture as our standard model.

\subsection{Regression Experiments on UCI Data}

We report the results of our regression experiments on the UCI datasets in
Table \ref{tbl:uci}. 
As in \citet{song2019distribution}, we evaluate the quality of forecasts using a check score $\rho_\tau(f,y) = \tau (y-f)$ if $y \geq f$ and $(1-\tau)(f-y)$ as in \citet{song2019distribution}; we average it over nine quantile levels $\tau \in {0.1,...,0.9}$.
We measure regression performance using the mean average error. 
These metrics precisely capture the two properties studied in our paper: the ability of our methods to ensure good calibration while having a very small regret relative to the baseline model as measured by a supervised learning loss.
% We evaluate Bayesian linear regression, Bayesian neural networks and compare against the baslines of \citet{kuleshov2018accurate} and \citet{song2019distribution}.

Our method improves over the accuracies and uncertainties of \citet{kuleshov2018accurate}, and in many cases over those of \citet{song2019distribution} on Bayesian linear regression, Bayesian neural networks, and deep ensembles, without ever being worse.
Note that also that out method is simpler and easier to implement than that of \citet{song2019distribution} (it does not require implementing variational inference), and applies to any input distribution, not just Gaussians.
% We report the latter for baseline and calibrated versions of Bayesian linear regression and the Bayesian neural network. We find that {\sc Calibration-By-Backprop} improves calibration error, although this improvement is variable, mainly because of numerical instabilities during training. Concrete dropout and deep ensembles provide better uncertainties than simple variational dropout or exact posterior inference, but their performance is below that of our model. 
% Recalibrated and base models achieve similar levels of mean average percent error. %forecasts achieve similar accuracies to the baselines, even though the latter have more parameters.% not suffer a drop in accuracy relative to the base forecaster.

% modified this to normal table as the wraptable env wasn't working properly here
\begin{table}
% \begin{wraptable}{r}{7cm}
% \vspace{-5mm}
\small
\begin{center}
\begin{tabular}{l|c|c|c}
\toprule
 & {\bf MNIST} & {\bf SVHN} & {\bf CIFAR10} \\
\midrule
\midrule
{\bf Base Model} &  &  &  \\
Accuracy & 0.9952 & 0.9508 & 0.9179 \\
Calibration & 0.3166 & 0.5975 & 0.5848 \\
\midrule
{\footnotesize\bf Platt Scaling} &  &  &  \\
Accuracy & 0.9952 & 0.9508 & 0.9181 \\
Calibration & 0.2212 &  0.3278 & 0.2233 \\
\midrule
{\bf Ours} &  &  &  \\
Accuracy & 0.9951 & 0.9509 & 0.9163 \\
Calibration & 0.1030 & 0.2674 & 0.1091 \\
% also have CIFAR10 backprop for smaller nnet with 0.9052 acc and 0.2034 calib score
\bottomrule
\end{tabular}
\end{center}
\caption{Performance on Image Classification\label{classification}}
% \vspace{-10mm}
% \end{wraptable}
\end{table}

\subsection{Classification Experiments on MNIST, SVHN, CIFAR10}

We report the results of the image classification experiments in
Table \ref{classification}. We measure performance using accuracy and calibration error of \citet{kuleshov2018accurate} on the test set.
We report these metrics for baseline and calibrated versions of convolutional neural network classifier. We perform recalibration with a simple softmax regression (a multi-class generalization of Platt scaling) and with the neural network recalibrator. The best uncertainties are produced by our method, while accuracies are similar.
%Recalibrated and base models achieve similar levels of accuracy.

%% file: discussion.tex
\section{Related Work}

\paragraph{Probabilistic Forecasting}

More modern discussions of probabilistic forecasting can be found in the literature on  meteorology \citep{gneiting2005weather}. This influential work appears in methods weather forecasting applications systems \citep{raftery2005using}. Most previous work focuses on classification, but recent work \citep{gneiting2007probabilistic,kuleshov2018accurate} extends to regression.

Probabilistic forecasting has been studied extensively in the statistics literature \citep{murphy1973vector,dawid1984prequential}, mainly in the context of evaluation using proper scoring rules \citep{gneiting2007strictly}. Proper scores measure calibration and sharpness in classification \citep{murphy1973vector} and regression \citep{hersbach2000decomposition}. 
% \citet{dawid1984prequential} study this problem in the context of Bayesian statistics.

\paragraph{Calibration}

Recalibration is a widely used approach for improving probabilistic forecasts. It has roots in the classification setting, starting with Platt scaling \citep{platt1999probabilistic} and isotonic regression \citep{niculescu2005predicting}. These mehtods have been extended to multi-class \citep{zadrozny2002transforming}, structured \citep{kuleshov2015calibrated}, and online prediction \citep{kuleshov2017estimating}.
There is significant recent interest in calibration in deep learning
\citep{guo2017calibration,lakshminarayanan2016simple,gal2017concrete,kuleshov2018accurate}.
Beyond deep learning, accurate uncertainty estimation is important for the design of decision-making systems
~\citep{malik2019calibrated}, crowdsourcing \citep{werling15adaptive}, data-efficient machine learning \citep{ratner2016data,kuleshov2017deep,ren2018learning}, machine translation \citep{kumar2019calibration}, as well as other problems in natural language processing and beyond \citep{nguyen2015posterior,kuleshov2019machine}.

Conformal prediction is a closely related line of work \citep{vovk2005algorithmic,shafer2008tutorial}.
The recent methods by \citet{romano2019conformalized} and \citet{barber2021predictive} provide similar guarantees, but only for a single confidence interval (i.e., two quantiles), not a full distribution.
\citet{romano2020classification} focuses on {\em classification}, while our work encompasses regression.

Compared to previous work, our methods target a different type of calibration (distribution vs. quantile) and estimate different distributions ($P(Y \leq F_X^{-1}(p))$ vs. $P(Y|H(X) = F)$) using different objectives (e.g., the quantile divergence vs.~calibration error in \citet{kuleshov2018accurate}).% \vk{remove?}

\paragraph{Distribution Calibration}

%We propose a new recalibration technique
Unlike \citet{song2019distribution} our method can recalibrate any parametric distribution (not just Gaussians) while being also simpler. While \citet{song2019distribution} relies on variational inference in Gaussian processes (which is slow and complex to implement), our method uses a neural network that can be implemented in a few lines of code.
Our method applies to both classification and regression and outperforms \citet{song2019distribution}, \citet{kuleshov2018accurate}, as well as Platt and temperature scaling.

Interestingly, the method of \citet{song2019distribution} is a special case of Algorithm \ref{alg:recal0} when $\phi(F)$ consists of Gaussian natural parameters, $R$ is a Gaussian process, its outputs are parameters for the Beta link function, and $L$ is the log-likelihood. However, the resulting problem requires variational inference; our framework instead admits simple solutions based on gradient descent.

\section{Conclusion}

Accurate predictive uncertainties are fully characterized by two properties---calibration and sharpness.
We argue that predictive uncertainties should maximize sharpness subject to being calibrated \citep{gneiting2007probabilistic}  and propose
a recalibration technique that achieves this goal.
Out technique guarantees distribution calibration for a wide range of base models and is easy to implement in a few lines of code.
It applies to both classification and regression and is formally guaranteed to produce asymptotically distributionally calibrated forecasts while minimizing regret. 
Finally, our analysis formalizes the well-known paradigm of \citet{gneiting2007probabilistic} and provides a simple method that provably achieves it. This lends strong support for this principle and influences how one should reason about uncertainty in machine learning.
We believe that an important takeaway of our work is that calibration should be leveraged more broadly throughout machine learning.